\documentclass[10pt,letterpaper]{article}
\usepackage[top=0.85in,left=1in,footskip=0.75in,marginparwidth=2in]{geometry}

\usepackage[utf8]{inputenc}

\usepackage{nameref,hyperref}

\usepackage{microtype}
\DisableLigatures[f]{encoding = *, family = * }

\raggedright
\usepackage{changepage}

\usepackage[aboveskip=1pt,labelfont=bf,labelsep=period,singlelinecheck=off]{caption}

\makeatletter
\renewcommand{\@biblabel}[1]{\quad#1.}
\makeatother

\usepackage{lastpage,fancyhdr,graphicx}
\usepackage{epstopdf}
\fancyheadoffset[L]{2.25in}
\fancyfootoffset[L]{2.25in}

\usepackage{color}

\definecolor{Gray}{gray}{.25}

\usepackage{graphicx}

\usepackage{sidecap}

\usepackage{wrapfig}
\usepackage[pscoord]{eso-pic}
\usepackage[fulladjust]{marginnote}
\usepackage{amsmath}
\reversemarginpar
\begin{document}
\vspace*{0.35in}

\begin{flushleft}
{\Large
\textbf\newline{Advancing Heatwave Forecasting via Distribution Informed-Graph Neural Networks (DI-GNNs): Integrating Extreme Value Theory with GNNs}
}
\newline
\\
Farrukh A. Chishtie\textsuperscript{1,5,7*},
Dominique Brunet\textsuperscript{2},
Rachel H. White\textsuperscript{3},
Daniel Michelson\textsuperscript{2},
Jing Jiang\textsuperscript{4},
Vicky Lucas\textsuperscript{1},
Emily Ruboonga\textsuperscript{1},
Sayana Imaash \textsuperscript{3},
Melissa Westland\textsuperscript{3},
Timothy Chui\textsuperscript{3},
Rana Usman Ali\textsuperscript{7},
Mujtaba Hassan\textsuperscript{6,7},
Roland Stull\textsuperscript{3},
David Hudak\textsuperscript{2}

\bigskip
\textsuperscript{1}Institute for Resources, Environment and Sustainability, University of British Columbia, Vancouver, BC, Canada
\\
\textsuperscript{2}Environment and Climate Change Canada, Toronto, Ontario, Canada
\\
\textsuperscript{3}Department of Earth, Ocean and Atmospheric Sciences, University of British Columbia, Vancouver, BC, Canada
\\
\textsuperscript{4}Department of Forestry, University of British Columbia, Vancouver, BC, Canada
\\
\textsuperscript{5}Department of Occupational Science and Occupational Therapy, University of British Columbia, Vancouver, BC, Canada
\\
\textsuperscript{6}Department of Space Science, Institute of Space Technology, Islamabad, Pakistan
\\
\textsuperscript{7}Peaceful Society, Science and Innovation Foundation. Vancouver, BC, Canada

\bigskip
*farrukh.chishtie@gmail.com

\end{flushleft}

\section*{Abstract}
Heatwaves, prolonged periods of extreme heat, have intensified in frequency and severity due to climate change, posing substantial risks to public health, ecosystems, and infrastructure. Despite advancements in Machine Learning (ML) modeling, accurate heatwave forecasting at weather scales (1–15 days) remains challenging due to the non-linear interactions between atmospheric drivers and the rarity of these extreme events. Traditional models relying on heuristic feature engineering often fail to generalize across diverse climates and capture the complexities of heatwave dynamics. This study introduces the Distribution-Informed Graph Neural Network (DI-GNN), a novel framework that integrates principles from Extreme Value Theory (EVT) into the graph neural network architecture. DI-GNN incorporates Generalized Pareto Distribution (GPD)-derived descriptors into the feature space, adjacency matrix, and loss function to enhance its sensitivity to rare heatwave occurrences. By prioritizing the tails of climatic distributions, DI-GNN addresses the limitations of existing methods, particularly in imbalanced datasets where traditional metrics like accuracy are misleading. Empirical evaluations using weather station data from British Columbia, Canada, demonstrate the superior performance of DI-GNN compared to baseline models. DI-GNN achieved significant improvements in balanced accuracy, recall, and precision, with high AUC and average precision scores, reflecting its robustness in distinguishing heatwave events. These advancements highlight DI-GNN's potential as a transformative tool for heatwave prediction, with broader applications to other climate extremes such as cold snaps, floods, droughts, and atmospheric blocking events. Beyond meteorological forecasting, the framework offers scalable solutions for disaster preparedness, infrastructure resilience, and climate adaptation strategies, enabling actionable insights for policymakers and emergency response teams.

\section{Introduction}

Heatwaves (HWs), defined as prolonged periods of extreme heat relative to local climatological norms, are among the most severe climate-related hazards globally. These events have increased in frequency, intensity, and duration over recent decades due to anthropogenic climate change \cite{barriopedro2023,ipcc2021}. The Pacific Northwest heatwave of June 2021 exemplifies their devastating impacts, breaking historical temperature records by over 45°C. In British Columbia alone, this event resulted in over 700 excess deaths, widespread agricultural losses, and significant ecological disruptions, including the destruction of marine habitats \cite{white2023,thompson2022}. Such events highlight the urgent need for reliable HW forecasting to mitigate socioeconomic and environmental consequences.

Accurate HW forecasting at weather lead times (1–15 days) is crucial but remains challenging due to the complex interplay of large-scale atmospheric circulation, regional feedbacks, and local meteorological phenomena \cite{barriopedro2023,miralles2014,findell2017}. Traditional methods, such as statistical modeling and heuristic feature engineering, are often insufficient for predicting rare and extreme HW events. They struggle to generalize across diverse climates and capture the nonlinear dynamics of HW formation \cite{perkins2015,mcgregor2015,vogel2020}.

Artificial Intelligence (AI) is revolutionizing the study of climate extremes, offering powerful tools for analyzing and forecasting these rare yet increasing frequent and impactful events. Techniques such as machine learning (ML) and deep learning have demonstrated remarkable success in understanding and predicting phenomena like droughts, marine heatwaves, and extreme rainfall events. For instance, ML models have been employed for subseasonal predictions of temperature extremes, leveraging their ability to uncover complex relationships between atmospheric drivers and surface anomalies at various time scales \cite{vanstraaten2022, weyn2021,lopezgomez2022}. Deep learning approaches like neural networks are increasingly being used to model land-atmosphere interactions and sea surface temperature anomalies, offering insights into the dynamical processes underlying heatwaves and other climate extremes \cite{rasp2020, scher2021, reichstein2019}. These data-driven approaches complement traditional dynamical models by improving computational efficiency and enabling high-resolution ensemble forecasts for extreme events.

Graph Neural Networks (GNNs) have emerged as a transformative tool in meteorology, enabling the modeling of complex spatial and temporal dependencies inherent in atmospheric phenomena. By representing meteorological data as graphs—where nodes correspond to observation points (e.g., weather stations or grid cells) and edges capture the relationships between these points—GNNs facilitate the analysis of intricate interactions within weather systems, leading to more accurate and efficient forecasts. A notable application of GNNs in weather forecasting is GraphCast, an AI model developed by DeepMind. GraphCast utilizes GNNs to process spatially structured data, allowing it to predict weather conditions up to 10 days in advance with high accuracy. It has demonstrated superior performance compared to traditional numerical weather prediction systems \cite{lam2022}. Another significant development is the Artificial Intelligence Forecasting System (AIFS) by the European Centre for Medium-Range Weather Forecasts (ECMWF). AIFS employs a GNN-based encoder and decoder, combined with a sliding window transformer processor, to produce medium-range global weather forecasts. Trained on ECMWF's ERA5 reanalysis and operational numerical weather prediction analyses, AIFS has shown high skill in forecasting upper-air variables, surface weather parameters, and tropical cyclone tracks \cite{dueben2023}. The FuXi model represents a cascade machine learning forecasting system for 15-day global weather forecasts. FuXi integrates GNNs with other machine learning techniques to enhance the accuracy of medium-range weather predictions \cite{chen2023}. These advancements underscore the potential of GNNs to revolutionize weather forecasting by providing faster and more precise predictions. By leveraging the relational structure of meteorological data, GNNs facilitate the modeling of complex atmospheric dynamics, paving the way for significant progress in climate science and meteorological research.

Among these AI/ML applications, heatwave prediction has gained significant attention due to the increasing frequency and intensity of these events under climate change. Statistical and AI-based models have advanced the ability to predict heatwaves across diverse temporal and spatial scales. Subseasonal prediction models, including random forests and linear regression, have demonstrated skill in forecasting extreme warm days by incorporating soil moisture deficits, atmospheric circulation patterns, and tropical teleconnections \cite{guigma2021, chattopadhyay2020,miller2021}. The introduction of neural networks for heatwave forecasting has further enhanced predictive accuracy, particularly when integrating predictors like sea ice and snow cover anomalies, as well as blocking patterns over the Northern Hemisphere \cite{vanstraaten2022,quinting2017,seneviratne2012}. GNNs also offer a promising approach to HW forecasting. GNNs leverage graph structures to model spatiotemporal dependencies, capturing both local and global relationships in meteorological data \cite{scarselli2009,kipf2017,moore2023}. For example, Li et al. \cite{li2023} demonstrated the utility of GNNs in regional HW prediction, yet such existing frameworks often rely on heuristic adjacency matrices and standard evaluation metrics, limiting their effectiveness in identifying rare extremes \cite{kipf2017,moore2023}. To address these gaps, this study introduces a novel framework: the Distribution Informed-Graph Neural Network (DI-GNN).

DI-GNN integrates Extreme Value Theory (EVT), a statistical framework designed to model the tails of distributions, making it uniquely suited to predict rare and extreme events \cite{coles2001,embrechts1997}. EVT-derived parameters, such as those from the Generalized Pareto Distribution (GPD), are incorporated into the GNN architecture, enhancing its ability to capture the statistical properties of HW extremes. This integration represents a significant departure from traditional feature engineering, embedding theoretical rigor into the forecasting process \cite{vogel2020,zscheischler2020,bastos2020}. More generally, we propose that other representations of statistical distributions be integrated with ML architectures in order to meaningfully improve model performance and insights for particular phenomena under study. 

Our empirical evaluations demonstrate that DI-GNN outperforms the baseline model of Li et al, achieving higher precision, recall, and balanced accuracy, particularly in forecasting HW events at weather lead times. By prioritizing the tails of climatic distributions, DI-GNN not only improves predictive accuracy but also provides a more interpretable framework for understanding the drivers of HWs. These advancements are particularly relevant for regions like British Columbia, where extreme HW events have profound public health, economic, and ecological implications \cite{white2023,thompson2022}.

This paper is structured to provide a comprehensive exploration of the intersection between climate science and machine learning, with a particular focus on advancing methodologies for heatwave prediction. We begin next with an introduction to Graph Neural Networks (GNNs), emphasizing their ability to model complex spatiotemporal dependencies in climate data. Building on this foundation, the next section examines the application of GNNs for spatiotemporal heatwave prediction, highlighting the challenges of forecasting rare and extreme weather events. The core contribution of this work, the Distribution-Informed GNN (DI-GNN), is then introduced. DI-GNN innovatively integrates Extreme Value Theory (EVT) to enhance the model's sensitivity to rare heatwave occurrences, offering a statistically rigorous approach to improving prediction accuracy.

Following this, the results section presents the experimental evaluation of DI-GNN, showcasing its superior performance in terms of metrics such as precision, recall, and balanced accuracy when compared to baseline models. A dedicated discussion explores how DI-GNN transcends traditional feature engineering, embedding theoretical rigor into every aspect of the modeling process. Finally, the paper concludes by summarizing the key contributions and outlining future directions, including extensions of DI-GNN to other climate extremes and its integration with alternative machine learning frameworks.

By bridging the gap between climate science and machine learning, DI-GNN offers a scalable and generalizable solution for real-time heatwave prediction. Beyond meteorological forecasting, its applications extend to disaster preparedness, infrastructure resilience, and climate adaptation strategies, providing timely and actionable insights for policymakers and emergency response teams.

\section{Graph Neural Networks (GNNs)}

Graph Neural Networks (GNNs) are a class of deep learning models specifically designed for graph-structured data. A graph \( G = (V, E) \) consists of nodes \( V \) and edges \( E \), where each node \( v_i \) is associated with a feature vector \( \mathbf{x}_i \). GNNs have demonstrated remarkable success in domains such as chemistry, social networks, and meteorology due to their ability to capture both local and global dependencies within graph data \cite{scarselli2009,kipf2017,velickovic2018graph}. 

In meteorological modeling, nodes represent weather stations, and edges encode spatial or statistical relationships based on geographic proximity or correlation in observed variables \cite{li2023,moore2023}. This flexible representation allows GNNs to model complex spatiotemporal dependencies, making them particularly well-suited for tasks like heatwave prediction, where interactions between weather stations are critical.

The key mechanism underlying GNNs is message passing, where information from neighboring nodes is aggregated to update a node's representation. At each layer \( k \), the representation of node \( v_i \) is updated as:
\begin{equation}
\mathbf{h}_i^{(k+1)} = \phi \left( \mathbf{h}_i^{(k)}, \sum_{j \in \mathcal{N}(i)} \alpha_{ij}^{(k)} W^{(k)} \mathbf{h}_j^{(k)} \right),
\end{equation}
where \( \mathcal{N}(i) \) denotes the neighbors of node \( i \), \( W^{(k)} \) is a learnable weight matrix, and \( \phi \) is a non-linear activation function. The attention coefficient \( \alpha_{ij}^{(k)} \), introduced in Graph Attention Networks (GATs) \cite{velickovic2018graph}, modulates the contribution of neighboring nodes and is computed as:
\begin{equation}
\alpha_{ij}^{(k)} = \frac{\exp\left( \text{LeakyReLU}\left( a^\top [W^{(k)} \mathbf{h}_i^{(k)} \| W^{(k)} \mathbf{h}_j^{(k)}] \right) \right)}{\sum_{l \in \mathcal{N}(i)} \exp\left( \text{LeakyReLU}\left( a^\top [W^{(k)} \mathbf{h}_i^{(k)} \| W^{(k)} \mathbf{h}_l^{(k)}] \right) \right)},
\end{equation}
where \( \| \) denotes concatenation, and \( a \) is a learnable parameter vector.

The Leaky Rectified Linear Unit (LeakyReLU) is an activation function used in this context to address the issue of so-called dying neurons, which occurs when certain nodes produce zero gradients during training. The function is defined as:
\begin{equation}
\text{LeakyReLU}(x) = 
\begin{cases} 
x, & \text{if } x \geq 0, \\
\alpha x, & \text{if } x < 0,
\end{cases}
\end{equation}
where \( \alpha \) is a small positive scalar (e.g., \( \alpha = 0.01 \)). This formulation allows a small gradient to flow when the input is negative, ensuring that all nodes contribute to the optimization process.

Through multiple layers of message passing, GNNs propagate information across the graph, enabling the model to learn node representations that incorporate both local features and the broader graph structure. Advanced GNN architectures, such as GATs and Graph Convolutional Networks (GCNs), further refine this process by introducing mechanisms like attention weighting and spectral filtering, respectively.\cite{kipf2017,velickovic2018graph} These innovations allow GNNs to dynamically adjust to the varying importance of neighboring nodes and edges.

\section{Spatiotemporal Heatwave Prediction using Graph Neural Networks}

\cite{li2023} developed a spatiotemporal Graph Neural Network (GNN) framework to predict heatwaves using meteorological data from 91 weather stations across the contiguous United States (CONUS) over a 15-year period (2006–2020). The framework integrated spatial relationships and temporal sequences to model heatwave dynamics, offering a novel approach to capturing the complex interplay of atmospheric variables driving extreme heat events.

Each weather station \( i \) was represented by a feature vector \( \mathbf{x}_i^{(t)} \), comprising meteorological and geographic variables:

\begin{equation}
\mathbf{x}_i^{(t)} = [\beta_i^{(t)},T_{\max}, T_{\min}, T_{\text{dew}}, \text{RH}, U, P, \text{Lon}, \text{Lat}, \text{DOY}, P_{sea}, \text{ONI}, T_{\text{thresh}}],
\end{equation}

where \( T_{\max} \) and \( T_{\min} \) are maximum and minimum daily temperatures, \( T_{\text{dew}} \) is dew point temperature, and \( \text{RH} \) is relative humidity. Other variables include wind speed (\( U \)), precipitation (\( P \)), geographic coordinates (\( \text{Lon}, \text{Lat} \)), day of the year (\( \text{DOY} \)), sea level pressure (\( {P_sea} \)), the Oceanic Niño Index (\( \text{ONI} \)), and \( T_{\text{thresh}} \), which are long-term normal temperatures. Heatwave events were classified using the Perkins-Kirkpatrick and Lewis (PKL) criteria \cite{perkins2015}. Such events occur if \( T_{\max} > T_{\text{90}} \) for three or more consecutive days during the summer period, ranging from May to September, whereby the threshold denotes the 90th percentile of daily maximum temperature. Heatwave days were labeled \( \beta_i^{(t)} = 1 \) during these events. 

Temporal sequences of input length \( C_{\text{in}} \) (e.g., the past 30 days) were used to predict heatwave occurrences over an output horizon \( C_{\text{out}} \) (e.g., the next 5 days):
\begin{equation}
\mathbf{X}_i^{(t)} = [\mathbf{x}_i^{(t-C_{\text{in}}+1)}, \mathbf{x}_i^{(t-C_{\text{in}}+2)}, \dots, \mathbf{x}_i^{(t)}].
\end{equation}
Spatial relationships were encoded through an adjacency matrix based on Pearson correlation coefficients, calculated as:
\begin{equation}
\rho_{ij} = \frac{\sum_{t=1}^T (x_i^t - \mu_i)(x_j^t - \mu_j)}{\sigma_i \sigma_j T},
\end{equation}
where \( \mu_i \) and \( \sigma_i \) are the mean and standard deviation of \( x_i^t \) over time \( T \). This approach captured the linear dependencies among weather stations, reflecting the spatial coherence of meteorological variables such as temperature and precipitation.

\subsection{Evaluation of Model Performance and Limitations of Standard Accuracy}

While Li et al.'s GNN framework achieved an impressive overall accuracy of 94.1\%, the results revealed significant limitations in its ability to predict rare heatwave events. Specifically, the model's recall (58.5\%) (also know as Probability of Detection) and precision (62.5\%) (also known as Success Rate) for heatwave occurrences highlighted a high rate of false negatives (missed heatwave days) and false positives (non-heatwave days predicted as heatwaves). These findings underscore the inadequacy of standard accuracy metrics for imbalanced datasets where the majority class dominates predictions. 

\subsection{Metrics for Model Evaluation}

Accuracy is traditionally calculated as:
\begin{equation}
\text{Accuracy} = \frac{\text{TP} + \text{TN}}{\text{TP} + \text{TN} + \text{FP} + \text{FN}},
\end{equation}
where True Positives (\( \text{TP} \)) and True Negatives (\( \text{TN} \)) represent correctly predicted heatwave and non-heatwave days, while False Positives (\( \text{FP} \)) and False Negatives (\( \text{FN} \)) correspond to misclassified events. However, for imbalanced datasets like those used in heatwave forecasting, accuracy is misleading. For instance, if heatwave days constitute only 12\% of the dataset, a naive model predicting "no heatwave" would achieve 88\% accuracy while failing completely at detecting actual heatwave days.

\textbf{Balanced Accuracy} provides a more equitable evaluation for imbalanced datasets:
\begin{equation}
\text{Balanced Accuracy} = \frac{\text{TPR} + \text{TNR}}{2},
\end{equation}
where \textbf{\( \text{TPR} \) (True Positive Rate, Recall, or also known as Probability of Detection (POD) in Weather Forecasting and Atmospheric Science communities)} measures the fraction of correctly identified heatwave days:
\begin{equation}
\text{Recall} = \frac{\text{TP}}{\text{TP} + \text{FN}},
\end{equation}
and \( \text{TNR} \) (True Negative Rate) quantifies the fraction of correctly identified non-heatwave days:
\begin{equation}
\text{TNR} = \frac{\text{TN}}{\text{TN} + \text{FP}}.
\end{equation}

\textbf{Precision (also known as Success Rate in Weather Forecasting and Atmospheric Science communities)} is another critical metric, defined as:
\begin{equation}
\text{Precision} = \frac{\text{TP}}{\text{TP} + \text{FP}},
\end{equation}
and reflects the accuracy of positive predictions. Low precision indicates a high number of false alarms, while low recall corresponds to missed heatwave events. 

The \textbf{F1 Score}, a harmonic mean of precision and recall, balances these two metrics:
\begin{equation}
\text{F1 Score} = 2 \cdot \frac{\text{Precision} \cdot \text{Recall}}{\text{Precision} + \text{Recall}}.
\end{equation}
The F1 score is particularly useful for imbalanced datasets, as it evaluates both precision and recall simultaneously, ensuring neither metric is disproportionately emphasized.

\textbf{Area Under the Curve (AUC)} for the Receiver Operating Characteristic (ROC) is a robust metric that evaluates the model's ability to distinguish between heatwave and non-heatwave days across all classification thresholds:
\begin{equation}
\text{AUC} = \int_{0}^{1} \text{TPR}(t) \, d\text{FPR}(t),
\end{equation}
where \( \text{FPR} \) is the False Positive Rate. A higher AUC indicates better model discrimination, with values closer to 1 reflecting a near-perfect separation between classes.

The \textbf{Precision-Recall Area Under Curve (PR-AUC)} evaluates performance on rare events by summarizing the precision-recall tradeoff across thresholds:
\begin{equation}
\text{PR-AUC} = \int_{0}^{1} \text{Precision}(r) \, dr,
\end{equation}
where \( r \) represents recall. Unlike AUC-ROC, PR-AUC focuses on performance for the minority class and is particularly informative when false negatives and false positives carry different costs.

\subsection{Implications for Heatwave Forecasting}

Metrics like balanced accuracy, precision, recall, F1 score, AUC, and PR-AUC provide a more nuanced evaluation of model performance for imbalanced datasets. By incorporating these metrics, models can effectively address limitations of conventional accuracy and ensure reliable detection of rare but critical heatwave events. Emphasizing these metrics is essential for minimizing missed detections and false alarms, thereby improving model applicability in real-world forecasting scenarios \cite{vogel2020,zscheischler2020}.

Moreover, the framework's reliance on heuristic methods for feature engineering and adjacency matrix construction also constrained its ability to generalize. The static Pearson-correlation-based adjacency matrix assumed fixed spatial relationships, failing to account for the dynamic nature of atmospheric processes, such as shifting jet streams or evolving El Niño-Southern Oscillation (ENSO) phases \cite{moore2023,velickovic2018graph}. Similarly, the use of a fixed 90th percentile threshold for \( T_{\text{max}} \) did not adapt to regional and seasonal variations in heat extremes, limiting its applicability across diverse climates \cite{perkins2015,mcgregor2015}.

These limitations underscore the need for dynamic graph structures that evolve over time to capture changing spatial relationships, as well as adaptive thresholds informed by statistical frameworks like Extreme Value Theory (EVT) \cite{coles2001,embrechts1997}. EVT-based approaches allow for more robust identification of extreme events by modeling the tails of temperature distributions, addressing the inadequacies of fixed percentile thresholds.

Incorporating advanced evaluation metrics, such as balanced accuracy and precision-recall curves, alongside dynamic and adaptive modeling techniques, would significantly enhance the robustness and interpretability of GNN-based heatwave prediction frameworks. These advancements would ensure more reliable forecasts, particularly for rare and impactful heatwave events, contributing to improved disaster preparedness and climate resilience strategies.

\section{Distribution Informed-GNN (DI-GNN)}

The DI-GNN framework builds upon and address key gaps in  the spatiotemporal GNN methodology established by \cite{li2023} using statistical theory of distributions, introducing the use of Generalized Pareto Distribution (GPD)-based insights into feature representation, adjacency matrix reweighting, and loss function design. By leveraging principles from Extreme Value Theory (EVT)\cite{coles2001,embrechts1997}, DI-GNN explicitly models the tail behavior of extreme events, enabling the framework to prioritize rare and impactful heatwave occurrences. 

The Generalized Pareto Distribution (GPD) is a cornerstone of extreme value theory, particularly well-suited for modeling the tails of distributions to predict rare and extreme events such as heatwaves. \cite{reich2014} demonstrated the efficacy of GPD within a hierarchical framework to account for serial dependencies in heatwave occurrences, particularly in the western United States, where temporal clustering of extreme temperatures is prominent. Similarly, \cite{mcgregor2024} emphasized the theoretical underpinnings of extreme value distributions, including the GPD, highlighting its utility in characterizing the intensity and frequency of heatwaves as statistically rare and physically impactful phenomena.

The DI-GNN framework integrates GPD parameters at multiple levels, embedding statistical rigor directly into the learning process and moving beyond traditional feature engineering approaches.

\subsection{GPD-Driven Feature Augmentation}

The GPD, a cornerstone of EVT, models the distribution of exceedances above a threshold \( u \). Its cumulative distribution function is:
\begin{equation}
H(y | \xi, \sigma) = 1 - \left( 1 + \frac{\xi y}{\sigma} \right)^{-1/\xi},
\end{equation}
where \( y = X - u \), \( \xi \) is the shape parameter, and \( \sigma \) is the scale parameter. These parameters capture the extremity of temperature distributions, with \( \xi > 0 \) representing heavy-tailed distributions, \( \xi = 0 \) denoting exponential decay, and \( \xi < 0 \) indicating bounded extremes \cite{coles2001}.

For each weather station \( i \), GPD parameters are fitted to historical maximum temperatures \( T_{\max} \), yielding a set of descriptors: \( \xi_i, \sigma_i, \mu_i, \text{Var}_i \) (variance), and \( q_{95} \) (95th percentile). These descriptors augment the input feature vector:
\begin{equation}
\tilde{\mathbf{x}}_i^{(t)} = [\mathbf{x}_i^{(t)}, \xi_i, \sigma_i, \mu_i, \text{Var}_i, q_{95}].
\end{equation}
This augmentation enables DI-GNN to incorporate the statistical properties of extreme values into the learning process, enhancing its sensitivity to rare and extreme events.

\subsection{Integration of GPD-Weighted Adjacency Matrices in GATs}

As represented in equation (1), GATs dynamically compute attention coefficients to determine the influence of neighboring nodes during message passing. The attention coefficients \(\alpha_{ij}^{(k)}\), introduced in Graph Attention Networks \cite{velickovic2018graph}, modulate the contribution of neighboring nodes to the updated representation. These coefficients are defined in Equation (2), which represents the concatenation of the transformed feature vectors of nodes \(i\) and \(j\), and \(\mathbf{a}\) is a learnable parameter vector.

To enhance GATs for heatwave forecasting or similar tasks, the integration of GPD-weighted adjacency matrices is proposed here. The GPD-weighted adjacency matrix, defined as:
\begin{equation}
A_{ij} = \rho_{ij} \cdot w_i \cdot w_j, \quad w_i = 1 + |\xi_i| + \frac{\sigma_i}{\max_j \sigma_j},
\end{equation}
introduces domain-specific statistical insights derived from Extreme Value Theory (EVT). Here, \(\rho_{ij}\) represents the Pearson correlation coefficient between nodes \(i\) and \(j\), while \(w_i\) incorporates GPD parameters such as the shape parameter \(\xi_i\) and the scale parameter \(\sigma_i\). These weights dynamically adjust the influence of nodes based on their statistical extremity.

In the context of GATs, the GPD-weighted adjacency matrix serves as a re-weighting mechanism for the attention coefficients \(\alpha_{ij}^{(k)}\). Specifically, the adjacency matrix can be used to initialize or mask the attention mechanism, ensuring that nodes with extreme characteristics, such as high-risk weather stations during heatwaves, exert a stronger influence on the learning process. This integration allows GATs to prioritize rare events or classes, effectively combining the flexibility of attention mechanisms with the statistical rigor of EVT.

By incorporating GPD weights into the adjacency structure, the GAT framework is extended to address limitations of static adjacency matrices, enabling dynamic adjustment of edge weights based on domain-specific characteristics. This approach not only enhances the interpretability of the learned attention coefficients but also improves the model's sensitivity to rare and extreme events, making it particularly suitable for applications in climate science, including extremes and for disaster prediction as well.

\subsection{Loss Function for Rare Event Detection}

To prioritize rare-event detection, DI-GNN introduces a GPD-weighted loss function. Weighted true positives (\( \text{TP}_{\text{weighted}} \)), false positives (\( \text{FP}_{\text{weighted}} \)), and false negatives (\( \text{FN}_{\text{weighted}} \)) are defined as:
\begin{equation}
\text{TP}_{\text{weighted}} = \sum_i w_i y_i \hat{y}_i, \quad \text{FP}_{\text{weighted}} = \sum_i w_i (1 - y_i) \hat{y}_i, \quad \text{FN}_{\text{weighted}} = \sum_i w_i y_i (1 - \hat{y}_i),
\end{equation}
where \( w_i = 1 + |\xi_i| + \frac{\sigma_i}{\max_j \sigma_j} \). The loss function is designed to maximize the weighted F1 score:
\begin{equation}
L_{\text{DI-GNN}} = 1 - \text{F1}_{\text{weighted}},
\end{equation}
with the weighted F1 score given by:
\begin{equation}
\text{F1}_{\text{weighted}} = \frac{(1 + \beta^2) \cdot \text{TP}_{\text{weighted}}}{(1 + \beta^2) \cdot \text{TP}_{\text{weighted}} + \beta^2 \cdot \text{FN}_{\text{weighted}} + \text{FP}_{\text{weighted}}}.
\end{equation}
The parameter \(\beta\) determines the relative importance of recall to precision, where \(\beta > 1\) prioritizes recall, \(\beta < 1\) prioritizes precision, and \(\beta = 1\) balances both equally. This approach ensures that the model emphasizes the correct classification of rare and extreme events, addressing imbalances in the dataset \cite{li2023}.

\subsection{Fitting GPD Parameters}

The process of fitting Generalized Pareto Distribution (GPD) parameters involved extracting key statistical features from \(T_{\text{max}}\) data for each weather station. First, the 90th percentile of \(T_{\text{max}}\) values is calculated as the threshold, above which exceedances were identified: 

\begin{equation}
\text{exceedances} = T_{\text{max}}[T_{\text{max}} > \text{threshold}] - \text{threshold}.
\end{equation}

Using these exceedances, the GPD parameters—shape (\(\xi\)) and scale (\(\sigma\))—were estimated via maximum likelihood estimation. In addition to \(\xi\) and \(\sigma\), other statistical descriptors such as the mean (\(\mu\)), variance (\(\sigma^2\)), and 95th percentile (\(q_{95}\)) were computed to capture the overall statistical properties of the temperature data. These parameters and descriptors are stored in a matrix, enabling the DI-GNN framework to incorporate distributional insights into its learning process. By including these statistical features, the model enhances its ability to detect and predict rare and extreme temperature events effectively.

\subsection{Evaluation Metrics}

To comprehensively assess DI-GNN's performance, evaluation metrics, besides precision, recall and F1 scores include:
\begin{itemize}
    \item Balanced Accuracy: Accounts for class imbalance by averaging recall for both heatwave and non-heatwave classes \cite{vogel2020}.
    \item AUC-ROC: Measures the model's discriminative ability \cite{velickovic2018graph}.
    \item Precision-Recall Curve and Average Precision (AP): Evaluate performance specifically on rare-event predictions.
\end{itemize}

In summary, the integration of EVT principles into the DI-GNN framework marks a significant departure from traditional feature engineering. By embedding GPD-derived parameters directly into the feature space, adjacency matrix, and loss function, DI-GNN leverages the statistical structure of extremes to enhance sensitivity to rare events. This theoretical foundation ensures that the model captures the probabilistic characteristics of heatwaves, enabling more accurate and interpretable predictions. Unlike heuristic methods, DI-GNN's EVT-based approach provides a principled and robust framework for forecasting extreme climate events \cite{coles2001,li2023}.

\section{Results}

\subsection{Model Setup and Data}    

Our study applied two Graph Neural Network (GNN)-based methodologies: the adaptation of Li et al's approach \cite{li2023} to British Columbia (BC), Canada, and the novel Distribution-Informed Graph Neural Network (DI-GNN) framework. In this study, we chose BC as an area of study, due its unique geography and climate as compared to other zones in this region. We further recognize that heatwaves depend on upper atmospheric dynamics and land-atmosphere interactions which are not included, as we are focused on lead weather scales. We intend to further investigate these in future studies to extend both spatial and lead time temporal scales by integrating related observables such soil moisture and land surface temperatures \cite{hu2023, zeppetello2024} from model, in-situ and satellite remote sensing sources.

Both methods leveraged data from 71 weather stations in BC from 2009 to 2024, processed from an hourly scale to daily from Environment and Climate Change Canada (ECCC). Meteorological variables, namely daily average (\( T_{avg} \)), maximum (\( T_{\max} \)) and minimum temperatures (\( T_{\min} \)), dew point temperature (\( T_{\text{dew}} \)), relative humidity (RH), precipitation (P), air pressure $P_{a}$, daily average wind speed (\(U\)), time and location of weather stations were used. These observations were processed into daily aggregates across stations to ensure compatibility. Heatwave Days were classified using the same PKL criteria as done in Li et al. \cite{li2023,perkins2015}. In the overall dataset, Heatwave Days represented 12.45\% of the entire dataset.  Since we were focused only on the weather-scale lead times, we did not include climate-related observables. Feature importance findings in \cite{li2023} also indicated that such observables contribute minimally to model performance at such short time scales. 

Figure 1 visualizes the spatial distribution of weather stations across BC, highlighting their geographical diversity.

\begin{figure}[ht!]
    \centering
    \includegraphics[width=0.8\textwidth]{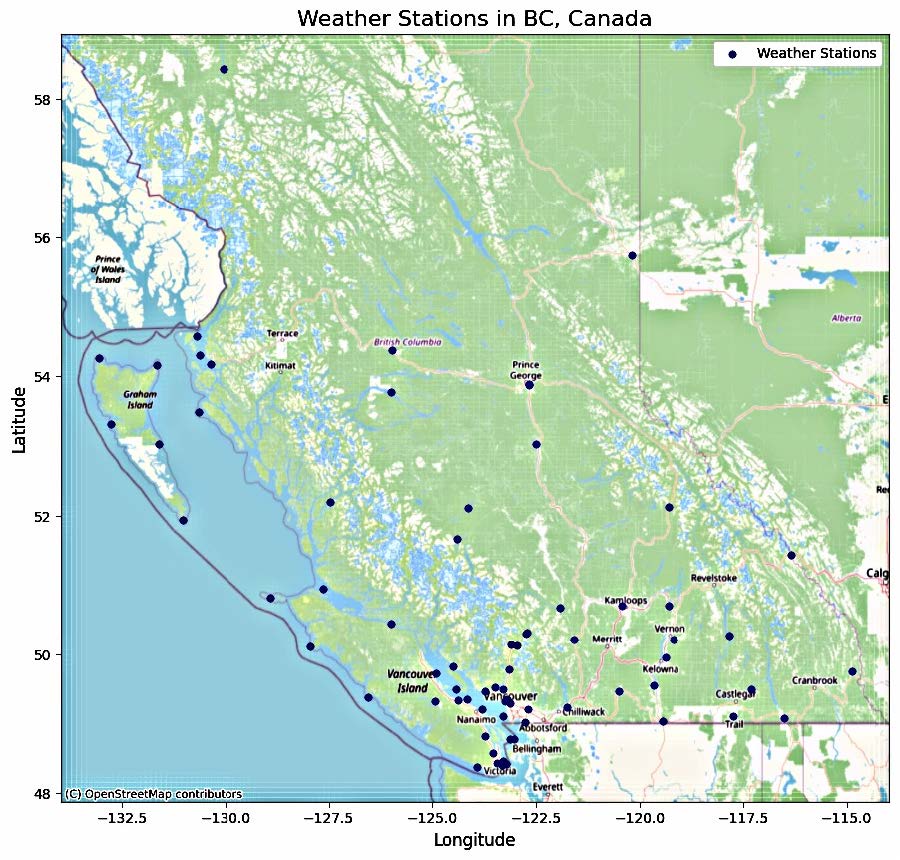}
    \caption{Spatial distribution of weather stations in British Columbia, Canada, used in the study. Data spans 71 stations from 2009 to 2024.}
\end{figure}

\subsection{Model Training and Experimental Setup}

Both models used historical weather data for training and validation, whereby the first 13 years were used for the former, and the last two years were for the latter. Two temporal configurations were evaluated: \( C_{\text{in}} = 10 \), \( C_{\text{out}} = 3 \) days and \( C_{\text{in}} = 10 \), \( C_{\text{out}} = 5 \) days, for a comparison of the DI-GNN model with the baseline model's optimal outputs. 

For DI-GNN, the Generalized Pareto Distribution (GPD) parameters (\( \xi \), \( \sigma \), mean, variance, and 95th percentile of \( T_{\max} \)) were derived from the training data and incorporated into the feature space, adjacency matrix, and loss function. Both models used a batch size of 64, Adam optimizer, a learning rate of \( 0.001 \), and early stopping based on validation loss. Training was performed on a high-performance GPU on the Google Colab to accelerate computations. For ML model implementation we used TensorFlow and Keras packages. 

\begin{figure}[ht!]
    \centering
    \includegraphics[width=0.9\textwidth]{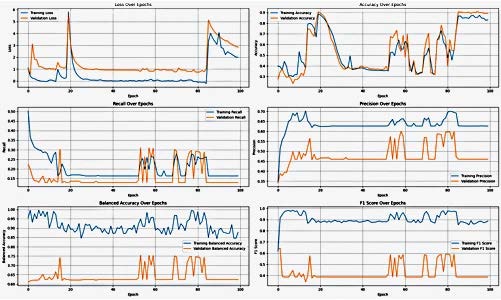}
    \caption{Performance metrics of the Li et al. (2023) model for heatwave prediction across 100 training epochs for the configuration \( C_{\text{in}} = 10 \) and \( C_{\text{out}} = 3 \). The metrics displayed include: Loss, Accuracy, Recall, Precision, Balanced Accuracy, and F1 Score, highlighting significant instability and low recall during training and validation phases.}
\end{figure}

\begin{figure}[ht!]
    \centering
    \includegraphics[width=0.9\textwidth]{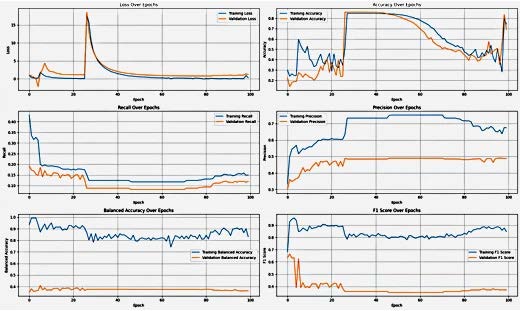}
    \caption{Performance metrics of the Li et al. (2023) model for heatwave prediction across 100 training epochs for the configuration \( C_{\text{in}} = 10 \) and \( C_{\text{out}} = 5 \). The metrics displayed include: Loss, Accuracy, Recall, Precision, Balanced Accuracy, and F1 Score, highlighting significant instability and low recall during training and validation phases.}
\end{figure}

\subsection{Baseline Results: Li et al.'s Model Adaptation for BC}

Li et al.'s GNN framework was adapted to BC's regional context using a similar architecture of Graph Attention Network (GAT) layers. Two configurations of temporal input (\( C_{\text{in}} \)) and output (\( C_{\text{out}} \)) horizons were tested: \( C_{\text{in}} = 10 \), \( C_{\text{out}} = 3 \) days, and \( C_{\text{in}} = 10 \), \( C_{\text{out}} = 5 \) days. Performance metrics are summarized in Table~\ref{tab:bc_baseline_results}.

\begin{table}[ht!]
\centering
\renewcommand{\arraystretch}{1.3} 
\setlength{\tabcolsep}{5pt} 
\caption{Performance metrics (BA = Balanced Accuracy, AUC = Area Under Curve) for the adaptation of Li et al.'s GNN model to BC weather station data.}
\label{tab:performance}
\label{tab:bc_baseline_results}
\footnotesize 
\begin{tabular}{|c|c|c|c|c|c|c|c|}
\hline
\textbf{Configuration} & \textbf{BA} & \textbf{Precision} & \textbf{Recall} & \textbf{F1 Score} & \textbf{Accuracy} & \textbf{AUC} & \textbf{Loss} \\ \hline
\( C_{\text{out}} = 3 \) & 62.35\% & 45.96\% & 12.86\% & 38.73\% & 89.15\% & 0.6750 & 0.8023 \\ \hline
\( C_{\text{out}} = 5 \) & 36.28\% & 48.94\% & 12.00\% & 37.25\% & 69.01\% & 0.5810 & 1.1945 \\ \hline
\end{tabular}
\end{table}

\begin{figure}[ht!]
    \centering
    \begin{minipage}[b]{0.45\textwidth}
        \centering
        \includegraphics[width=\textwidth]{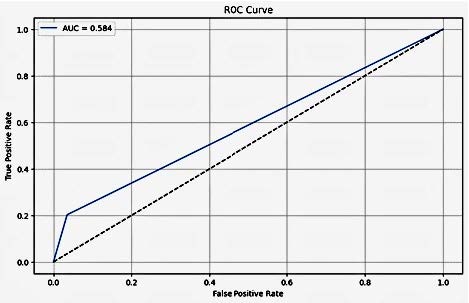}
        \caption*{\small (a) ROC Curve for \( C_{\text{in}} = 10 \), \( C_{\text{out}} = 3 \) (AUC = 0.584).}
    \end{minipage}
    \hfill
    \begin{minipage}[b]{0.45\textwidth}
        \centering
        \includegraphics[width=\textwidth]{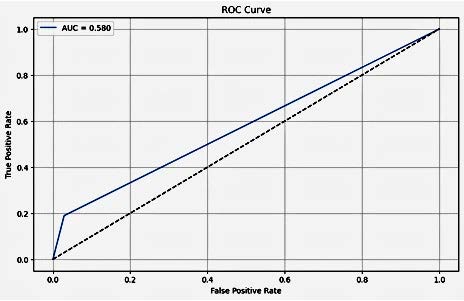}
        \caption*{\small (b) ROC Curve for \( C_{\text{in}} = 10 \), \( C_{\text{out}} = 5 \) (AUC = 0.580).}
    \end{minipage}
    \caption{ROC curves for the Li et al. model across two configurations. (a) Configuration \( C_{\text{in}} = 10 \), \( C_{\text{out}} = 3 \) shows an AUC of 0.584. (b) Configuration \( C_{\text{in}} = 10 \), \( C_{\text{out}} = 5 \) shows an AUC of 0.580.}
\end{figure}

While the adapted model achieved an overall accuracy of 89.15\% for \( C_{\text{out}} = 3 \), its recall and precision remained low due to class imbalance, highlighting its inability to effectively detect rare heatwave events. Figures 2 and 4 indicate significant instability and low recall during training and validation phases across all epochs. Both low AUC and Average Precision (AP) values shown in Figures 4 and 5 highlight the model's limited ability to distinguish heatwave events from non-heatwave days. These results underline the limitations of standard accuracy metrics and the need for advanced methodologies like DI-GNN.

\subsection{Enhanced Results with DI-GNN}

The DI-GNN framework incorporates GPD-based descriptors into the feature space, adjacency matrix, and loss function, explicitly modeling extreme events. Table~\ref{tab:di_gnn_results} summarizes the performance metrics for the DI-GNN under the same configurations.

\begin{table}[!ht]
\centering
\renewcommand{\arraystretch}{1.3} 
\setlength{\tabcolsep}{5pt} 
\caption{Performance metrics for the DI-GNN model (BA = Balanced Accuracy, AUC = Area Under Curve), incorporating GPD-derived parameters.}
\label{tab:di_gnn_results}
\footnotesize 
\begin{tabular}{|c|c|c|c|c|c|c|c|}
\hline
\textbf{Configuration} & \textbf{BA} & \textbf{Precision} & \textbf{Recall} & \textbf{F1 Score} & \textbf{Accuracy} & \textbf{AUC} & \textbf{Loss} \\ \hline
\( C_{\text{out}} = 3 \) & 86.66\% & 87.38\% & 84.95\% & 84.36\% & 43.77\% & 0.9190 & 0.1550 \\ \hline
\( C_{\text{out}} = 5 \) & 86.06\% & 85.26\% & 85.38\% & 83.72\% & 29.90\% & 0.9140 & 0.1443 \\ \hline
\end{tabular}
\end{table}

\begin{figure}[ht!]
    \centering
    \begin{minipage}[b]{0.45\textwidth}
        \centering
        \includegraphics[width=\textwidth]{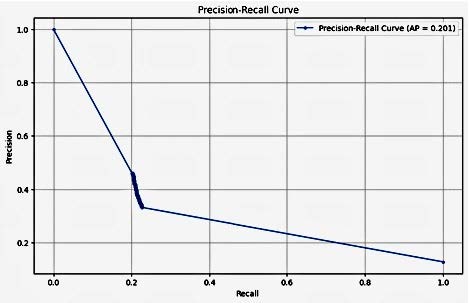}
        \caption*{\small (a) Precision-Recall Curve for \( C_{\text{in}} = 10 \), \( C_{\text{out}} = 3 \) (AP = 0.201).}
    \end{minipage}
    \hfill
    \begin{minipage}[b]{0.45\textwidth}
        \centering
        \includegraphics[width=\textwidth]{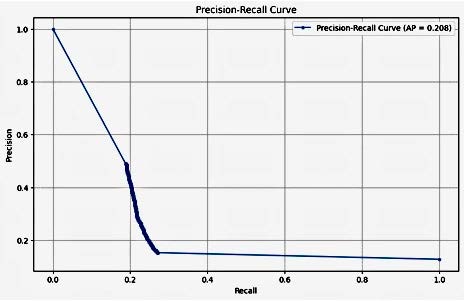}
        \caption*{\small (b) Precision-Recall Curve for \( C_{\text{in}} = 10 \), \( C_{\text{out}} = 5 \) (AP = 0.208).}
    \end{minipage}
    \caption{Precision-Recall curves for the Li et al. model across two configurations. (a) Configuration \( C_{\text{in}} = 10 \), \( C_{\text{out}} = 3 \) achieves an AP of 0.201, while (b) Configuration \( C_{\text{in}} = 10 \), \( C_{\text{out}} = 5 \) achieves an AP of 0.208. These low scores reflect the model's limited ability to correctly identify true positive heatwave days, leading to missed heatwave events in prediction.}
\end{figure}

DI-GNN achieved a balanced accuracy of 86.66\% for \( C_{\text{out}} = 3 \), significantly outperforming the baseline in recall (84.95\%) and precision (87.38\%). The high AUC of 0.919 and 0.914 respectively, highlights its robustness in distinguishing heatwave days from non-heatwave days. Both high AUC and Average Precision (AP) values shown in Figures 7 and 8 highlight the model's greater capability to distinguish heatwave events from non-heatwave days. These results highlight the significantly better results from the DI-GNN model.

\begin{figure}[!ht]
    \centering
    \includegraphics[width=0.9\textwidth]{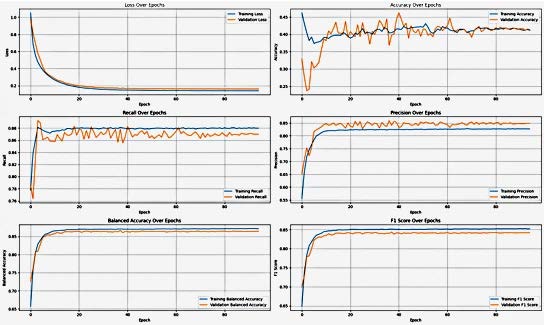}
    \caption{Performance metrics of the DI-GNN model for heatwave prediction across 100 training epochs for the configuration \( C_{\text{in}} = 10 \) and \( C_{\text{out}} = 3 \). Metrics displayed include: Loss, Accuracy, Recall, Precision, Balanced Accuracy, and F1 Score. DI-GNN demonstrates faster convergence, high recall, and consistent performance during training and validation phases.}
\end{figure}

\begin{figure}[!ht]
    \centering
    \includegraphics[width=0.9\textwidth]{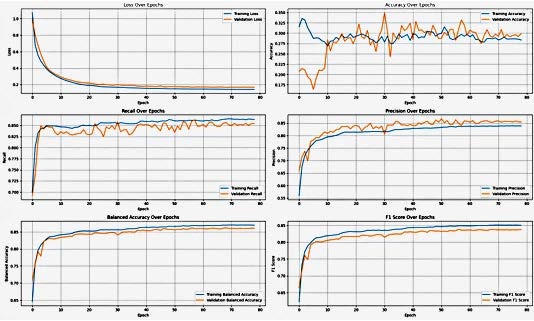}
    \caption{Performance metrics of the DI-GNN model for heatwave prediction across 100 training epochs for the configuration \( C_{\text{in}} = 10 \) and \( C_{\text{out}} = 5 \). Metrics displayed include: Loss, Accuracy, Recall, Precision, Balanced Accuracy, and F1 Score. The DI-GNN model demonstrates stable performance with improved recall and precision compared to baseline methods.}
\end{figure}

Importantly, the DI-GNN maintained stable performance across both configurations, demonstrating its adaptability to different temporal horizons. Figures 5 and 6 as compared to Figures 2 and 3, indicate significant stability and convergence in training and validation phases across all epochs. In comparing the results of the Li et al. framework \cite{li2023} and the Distribution-Informed Graph Neural Network (DI-GNN) for the \(C_{\text{in}}=10\) and \(C_{\text{out}}=3 \text{ and } 5\) days prediction cases, a clear distinction emerges in terms of loss trends and model robustness. As shown in the figures, the loss function in Li et al.'s model demonstrates significant instability during training, with sharp oscillations and slower convergence. This suggests a sensitivity to hyperparameter selection and an increased risk of overfitting, necessitating extensive model tuning to stabilize performance. Moreover, while Li et al.'s model achieves reasonable results for metrics such as recall, precision, balanced accuracy, and F1 score, these metrics fluctuate noticeably across epochs, highlighting potential issues with generalization and optimization stability.

On the other hand, DI-GNN exhibits a smoother and more stable loss convergence trend throughout training, underscoring its inherent statistical robustness. By leveraging Generalized Pareto Distribution (GPD)-based insights to reweight adjacency matrices and inform feature representations, DI-GNN effectively addresses the challenges associated with imbalanced datasets and rare-event prediction. This statistical grounding not only reduces the need for extensive hyperparameter tuning but also ensures consistent and reliable performance across all evaluated metrics. Thus, DI-GNN naturally integrates statistical reasoning to achieve state-of-the-art results, overcoming the limitations of manual tuning and instability observed in Li et al.'s framework.

\subsection{Comparative analysis and implications}

The results underscore the higher level of performance of DI-GNN in predicting rare heatwave events compared to the baseline model. By incorporating EVT principles, DI-GNN effectively addressed the class imbalance inherent in heatwave forecasting tasks, achieving higher recall and precision while minimizing false alarms. The improvement in balanced accuracy and F1 scores further validates the robustness of DI-GNN for operational use in disaster preparedness and climate adaptation strategies.

Figures 8 and 9 illustrate the Receiver Operating Characteristic (ROC) and Precision-Recall (PR) curves for both models, emphasizing the improved predictive performance of DI-GNN.

\begin{figure}[ht]
    \centering
    \begin{minipage}[b]{0.45\textwidth}
        \centering
        \includegraphics[width=\textwidth]{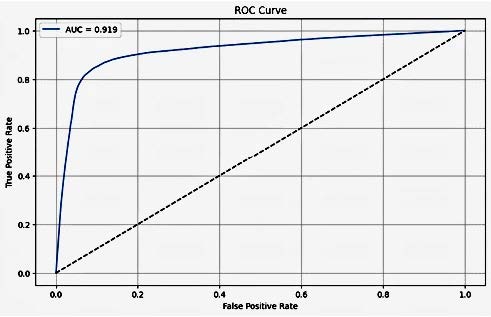}
        \caption*{\small (a) ROC Curve for \( C_{\text{in}} = 10 \), \( C_{\text{out}} = 3 \) (AUC = 0.919).}
    \end{minipage}
    \hfill
    \begin{minipage}[b]{0.45\textwidth}
        \centering
        \includegraphics[width=\textwidth]{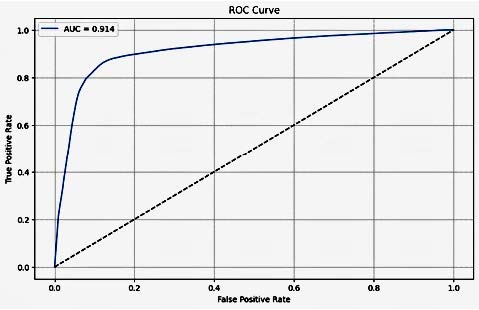}
        \caption*{\small (b) ROC Curve for \( C_{\text{in}} = 10 \), \( C_{\text{out}} = 5 \) (AUC = 0.914).}
    \end{minipage}
    \caption{ROC curves for the DI-GNN model across two configurations. (a) Configuration \( C_{\text{in}} = 10 \), \( C_{\text{out}} = 3 \) achieves an AUC of 0.919, reflecting superior discriminative ability. (b) Configuration \( C_{\text{in}} = 10 \), \( C_{\text{out}} = 5 \) achieves an AUC of 0.914. These results highlight DI-GNN's effectiveness in distinguishing heatwave events from non-heatwave days, significantly outperforming baseline models.}
\end{figure}

\begin{figure}[ht]
    \centering
    \begin{minipage}[b]{0.45\textwidth}
        \centering
        \includegraphics[width=\textwidth]{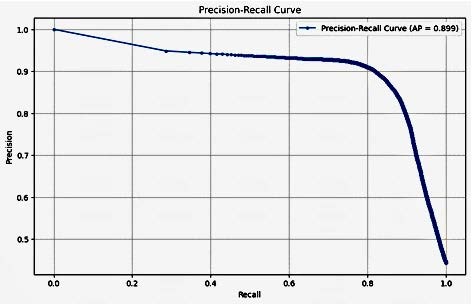}
        \caption*{\small (a) Precision-Recall Curve for \( C_{\text{in}} = 10 \), \( C_{\text{out}} = 3 \) (AP = 0.899).}
    \end{minipage}
    \hfill
    \begin{minipage}[b]{0.45\textwidth}
        \centering
        \includegraphics[width=\textwidth]{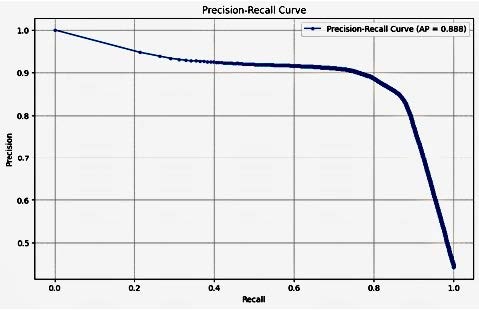}
        \caption*{\small (b) Precision-Recall Curve for \( C_{\text{in}} = 10 \), \( C_{\text{out}} = 5 \) (AP = 0.888).}
    \end{minipage}
    \caption{Precision-Recall curves for the DI-GNN model across two configurations. (a) Configuration \( C_{\text{in}} = 10 \), \( C_{\text{out}} = 3 \) achieves an AP of 0.899, while (b) Configuration \( C_{\text{in}} = 10 \), \( C_{\text{out}} = 5 \) achieves an AP of 0.888. These high AP scores reflect DI-GNN's superior ability to identify true positive heatwave days while minimizing false positives, effectively addressing the challenges of class imbalance and rare-event prediction.}
\end{figure}

\section{Discussion}

The Distribution-Informed Graph Neural Network (DI-GNN) represents a significant advancement in the modeling of rare and extreme events, transcending the limitations of traditional feature engineering by embedding Extreme Value Theory (EVT) principles directly into the learning process. This approach provides a more robust framework for predicting rare events like heatwaves, with several key distinctions and advantages over heuristic methods\cite{coles2001,embrechts1997,li2023}. 

Feature engineering is the process of transforming raw data into meaningful input variables for machine learning models. This involves the creation, selection, or transformation of features that are believed to improve the predictive performance of the model. Traditionally, feature engineering has been guided by domain expertise and involves techniques such as normalization, polynomial transformations, and manually derived metrics \cite{domingos2012few}. However, while effective for certain applications, feature engineering often lacks a formal statistical foundation, particularly in domains involving rare events or extreme values. This reliance on heuristics can lead to biases and diminished model generalizability. In the context of heatwave prediction, manually selecting features may overlook the tail behavior of the distribution, which is critical for modeling rare climate extremes.

\subsection{Advantages over traditional Feature Engineering}

Traditional feature engineering often relies on ad hoc or heuristic transformations, which may enhance input data but fail to rigorously capture the underlying statistical properties of extreme events \cite{perkins2015,mcgregor2015}. In contrast, DI-GNN leverages EVT to derive statistical descriptors specifically tailored to model extremes. These descriptors:
\begin{itemize}
    \item Depend on the distributional properties of extremes, ensuring relevance to rare-event modeling \cite{coles2001}.
    \item Influence multiple aspects of the modeling process, including graph structure, feature representation, and loss computation, embedding extreme value statistics throughout the framework \cite{li2023}.
    \item Ensure theoretical rigor by grounding the model in the probabilistic framework of EVT, providing a principled approach to modeling extremes \cite{embrechts1997}.
\end{itemize}

By using EVT-derived parameters, DI-GNN captures the tail behavior of the underlying distribution, providing a statistically robust alternative to the manual selection or transformation of features often used in traditional approaches.

\subsection{Comparative Analysis: Feature Engineering vs. DI-GNN}

The differences between feature engineering and DI-GNN can be summarized in Table~\ref{tab:comparison}, highlighting the theoretical and practical benefits of the latter.

\begin{table}[ht]
\centering
\renewcommand{\arraystretch}{1.3} 
\setlength{\tabcolsep}{8pt} 
\caption{Comparison between feature engineering and DI-GNN.}
\label{tab:comparison}
\footnotesize 
\begin{tabular}{|c|c|c|}
\hline
\textbf{Aspect} & \textbf{Feature Engineering} & \textbf{DI-GNN (EVT-Based)} \\ \hline
Origin & Ad hoc or heuristic & Statistically derived from EVT \\ \hline
Goal & Enhance input data & Integrate distributional properties \\ \hline
Dependence on Model & Independent of learning process & Shapes adjacency, features, and loss \\ \hline
Mathematical Basis & Lacks probabilistic foundation & Grounded in extreme value theory \\ \hline
Focus & Transform or select variables & Model the statistical nature of extremes \\ \hline
\end{tabular}
\end{table}

This comparison underscores DI-GNN’s emphasis on integrating theoretical insights into every layer of the learning process, moving beyond superficial enhancements to achieve a more holistic and statistically grounded framework.

\subsection{Statistical Robustness and Stability of DI-GNN}

A key virtue of the Distribution-Informed Graph Neural Network (DI-GNN) model lies in its inherent statistical robustness and stability during training and evaluation. Unlike traditional graph-based models such as Li et al.'s framework, which require extensive hyperparameter tuning to stabilize loss trends and optimize performance metrics, DI-GNN seamlessly integrates statistical principles derived from Extreme Value Theory (EVT) into its architecture. This integration significantly enhances its ability to generalize across diverse datasets and prediction horizons.

As demonstrated in the comparative results, the training loss for DI-GNN converges smoothly, avoiding the oscillatory behavior and instability observed in Li et al.'s model. The use of Generalized Pareto Distribution (GPD)-weighted adjacency matrices and features ensures that DI-GNN is not only tailored to predict rare and extreme events, such as heatwaves, but also robustly incorporates the tail characteristics of statistical distributions. This statistically grounded framework eliminates the reliance on heuristic adjustments, often required in conventional approaches, and automatically adapts to the underlying data distributions.

Moreover, DI-GNN's consistent performance across metrics such as recall, precision, balanced accuracy, and F1 score underscores its reliability in addressing the challenges posed by imbalanced datasets. While Li et al.'s model may exhibit good accuracy, this and other metrics' instability across epochs points to limitations in capturing the complexities of rare-event prediction. In contrast, DI-GNN naturally balances the trade-offs between sensitivity and specificity, making it an effective tool for both researchers and practitioners tackling climate extremes and other rare phenomena.

By embedding statistical reasoning directly into its learning process, DI-GNN transcends the need for manual hyperparameter tuning, offering a scalable and adaptable solution for real-world applications. Its virtues extend beyond heatwave forecasting, making it a robust model for other extreme weather events and imbalanced classification problems in environmental sciences.

\subsection{Implications for Rare-Event Prediction}

By embedding EVT-derived parameters into the graph structure, feature representation, and loss function, DI-GNN demonstrates how theoretical rigor can be effectively combined with machine learning to address real-world challenges\cite{coles2001,li2023}. The use of Generalized Pareto Distribution (GPD) parameters, such as shape (\( \xi \)) and scale (\( \sigma \)), ensures that the framework prioritizes the prediction of rare and impactful events. Additionally, the reweighting of the adjacency matrix and the introduction of a GPD-weighted loss function further enhance DI-GNN's ability to detect these rare occurrences, particularly in imbalanced datasets\cite{vogel2020,zscheischler2020}.

\subsection{Broader Implications}

The DI-GNN framework represents a significant advancement over traditional feature engineering approaches by embedding statistical distributional properties directly into the machine learning pipeline. Traditional feature engineering often relies on heuristic or ad hoc transformations of data, such as normalization or manual creation of domain-specific features, which can be effective but lack a formal statistical foundation. DI-GNN addresses this limitation by integrating principles from Extreme Value Theory (EVT), allowing for the systematic representation of rare and extreme events, such as heatwaves.

More generally, DI-GNN demonstrates the potential of incorporating representations of statistical distributions, such as those derived from EVT, into machine learning architectures to enhance both model performance and interpretability. By embedding statistical descriptors—such as the shape (\(\xi\)) and scale (\(\sigma\)) parameters of the Generalized Pareto Distribution (GPD)—into the feature space, adjacency matrices, and loss functions, DI-GNN explicitly models the tails of data distributions. This approach not only improves sensitivity to rare phenomena but also provides a principled framework for understanding the underlying drivers of these events.

Building on this idea, we propose that other representations of statistical distributions, such as those from Gaussian Mixture Models, Copulas, or Bayesian hierarchical models, be similarly integrated into machine learning architectures. Such integrations would enable models to better capture variability, dependence structures, and uncertainties inherent in the data. For example, these representations could be valuable in applications involving compound climate extremes, such as concurrent heatwaves and droughts, where joint statistical distributions are essential for accurate prediction.

The broader implication of this work lies in its generalizability. DI-GNN offers a blueprint for embedding statistical rigor into machine learning models, extending beyond heatwaves to other phenomena such as floods, hurricanes, or even non-climatic extremes in fields like finance or epidemiology. Furthermore, the flexibility of this approach suggests compatibility with other machine learning frameworks, including Convolutional Neural Networks (CNNs), Recurrent Neural Networks (RNNs), and Transformers, where the integration of statistical distributions could enrich spatiotemporal modeling and sequence prediction.

By moving beyond heuristic-driven approaches, DI-GNN exemplifies a shift toward hybrid models that combine statistical theory with machine learning to address domain-specific challenges. This integration ensures not only enhanced predictive accuracy but also deeper insights into the mechanisms underlying extreme events, offering actionable intelligence for policymakers, emergency planners, and researchers alike.

\section{Conclusion and Future Directions}

DI-GNN transcends traditional feature engineering by integrating the full statistical power of EVT into its framework. This innovation ensures that rare and extreme events are modeled with greater accuracy and theoretical rigor. By leveraging EVT-derived parameters for feature augmentation, adjacency re-weighting, and loss computation, DI-GNN provides a robust solution for rare-event prediction. Its ability to embed statistical insights directly into the ML modeling process positions it as a transformative approach to addressing challenges in climate resilience and disaster preparedness\cite{li2023,vogel2020}.

The integration of EVT into GNN frameworks like DI-GNN opens promising avenues for further research, as our model has several limitations. This includes sparsity of observations from limited weather stations. In our future work, we intend to explore:

\begin{itemize}
    \item Expanding the use of dynamic graph representations to account for including evolving spatial and temporal dependencies in climate data and upper atmosphere model output to extend model performance to subseasonal to seasonal scales \cite{moore2023,velickovic2018graph}.
    \item Applying DI-GNN to other extreme weather phenomena, such as cold extremes, floods or hurricanes, to evaluate its generalizability \cite{li2023}.
    \item Investigating the incorporation of additional EVT-based metrics, such as return levels or exceedance probabilities, to further enhance predictive capabilities \cite{coles2001}.
    \item Extending the spatial scale of the present model to across Canada along with comparison with Numerical Weather Prediction (NWP) models such as Global Deterministic Prediction System (GDPS) as well as ML-based models such as GraphCast, AIFS and Fuxi . 
    \item Inclusion of satellite remote sensing data such as Land Surface Temperature (LST) and Soil Moisture data to model performance to longer lead times. 
    
\end{itemize}

 With present findings, we propose that DI-GNN does offer unique advantages: it does not rely on complex data assimilation and is comparatively easier to set up. However, its current limitations, such as predictions constrained to station locations, highlight the need for trade-offs between simplicity and spatial coverage. Nonetheless, DI-GNN represents a complementary or supplementary approach for forecasters, capable of providing localized, actionable insights to complement global-scale systems.

Moving forward, the DI-GNN framework holds significant potential for further development and application. Incorporating dynamic graph representations to account for evolving spatial and temporal dependencies could extend its performance to subseasonal and seasonal time scales \cite{moore2023,velickovic2018graph}. Broadening the scope of DI-GNN to address other extreme weather events, including cold spells, floods, and hurricanes, would demonstrate its versatility \cite{li2023}. The inclusion of additional EVT-based metrics, such as return levels and exceedance probabilities, could deepen the model’s predictive capabilities \cite{coles2001}. Expanding its spatial application across Canada and comparing it directly to models like GDPS and GraphCast would provide valuable benchmarks. Furthermore, integrating satellite-derived data, such as Land Surface Temperature (LST) and soil moisture, could enhance predictive accuracy for regions with sparse observational coverage. Together, these advancements could position DI-GNN as a robust tool for climate resilience and disaster mitigation, capable of bridging the gap between operational forecasting and cutting-edge machine learning.

Future research will also aim to explore the integration of DI-GNN with other machine learning architectures, including Convolutional Neural Networks (CNNs), Recurrent Neural Networks (RNNs), and Transformers, to enhance predictive performance for diverse spatiotemporal datasets \cite{reichstein2019, rasp2020}. CNNs and Transformers, for instance, offer powerful tools for learning spatial and temporal patterns in climate data, while RNNs can capture sequential dependencies in meteorological time series. Moreover, combining dynamical model outputs with the statistical rigor of EVT presents an exciting opportunity to build hybrid models and frameworks including involving physics model outputs capable of leveraging the strengths of each approach. These efforts will not only improve the accuracy of rare-event forecasting but also advance our understanding of the physical mechanisms driving extreme climate events, creating pathways for robust and scalable AI-based climate solutions.

In summary, the DI-GNN framework, as demonstrated in this study, represents a robust advancement in addressing the challenges of heatwave prediction. By incorporating principles from Extreme Value Theory (EVT), DI-GNN effectively models the tails of climate data distributions, prioritizing the detection of rare and extreme events. Moreover, and more broadly, it can also incorporate other kinds of statistical distributions besides EVT. The success of DI-GNN highlights its applicability beyond heatwaves to other climate extremes, such as floods, droughts, and atmospheric blocking events, where capturing the statistical properties of extremes is equally critical. By adapting EVT-derived parameters to specific phenomena, DI-GNN can be extended to provide accurate, interpretable predictions for compound events, such as heatwave-induced wildfires or flood-drought sequences, thereby supporting disaster risk reduction, climate adaptation and resilience strategies. \cite{seneviratne2012, bastos2020}. 

\section*{Acknowledgments}
We thank the University of British Columbia (UBC) Climate Scholars program for funding this research which is part of an ongoing collaboration between UBC and ECCC. FAC would like to thank Peaceful Society, Science and Innovation Foundation along with Vancouver Foundation for financial support, including resources for high-performance computing.  

\section*{Code Availability}
Code and data is available upon request.

\end{document}